%% file: main.tex
\documentclass{article} 
\usepackage{iclr2019_conference,times}

\input{math_commands.tex}

\input{our-macro.tex}

\usepackage{hyperref}
\usepackage{url}
\usepackage{graphicx}
\usepackage{amssymb}
\usepackage{comment}
\usepackage{caption}
\usepackage{subcaption}
\usepackage{booktabs}

\usepackage[export]{adjustbox}

\title{Phrase-Based Attentions}

\author{Phi Xuan Nguyen, Shafiq Joty \\ 
Nanyang Technological University, Singapore\\
\texttt{xuanphi001@e.ntu.edu.sg, srjoty@ntu.edu.sg} \\
}

 \iclrfinalcopy 

\begin{document}

\maketitle

\begin{abstract}

Most state-of-the-art neural machine translation systems, despite being different in architectural skeletons (\eg\ recurrence, convolutional), share an indispensable feature: the Attention. However, most existing attention methods are token-based and ignore the importance of phrasal alignments, the key ingredient for the success of phrase-based statistical machine translation. In this paper, we propose novel phrase-based attention methods to model n-grams of tokens as attention entities. We incorporate our phrase-based attentions into the  recently proposed Transformer network, and demonstrate that our approach 
yields improvements of 1.3 BLEU for English-to-German and 0.5 BLEU for German-to-English translation tasks on WMT newstest2014 using WMT'16 training data.


\end{abstract}

\section{Introduction} \label{sec:intro}

\input{introduction.tex}
\section{Background} 
\input{background.tex}

\section{Multi-head Phrase-based Attention} \label{sec:phrase_attention}
\input{phrase_based_attention.tex}

\section{Experiments} \label{sec:experiments}
\input{experiments.tex}

\section{Conclusions} \label{sec:conclusion}

We have presented novel approaches to incorporating phrasal alignments into the attention mechanism of state-of-the-art neural machine translation models. Our methods assign attentions to all four possible mapping relations between target and source sequences: token-to-token, token-to-phrase, phrase-to-phrase and phrase-to-phrase. While we have applied our attention mechanisms to the Transformer network, they are generic and can be implemented in other architectures.
On WMT'14 English-to-German and German-to-English translation tasks, all of our methods surpass the Transformer base model. Our model with interleaved heterogeneous attention, which tackles all the possible phrasal mappings in a unified framework, achieves improvements of 1.33 BLEU for English-German and 0.48 BLEU for German-English translation tasks over the baseline model. We are planning future extensions of our techniques to other tasks, such as summarization and question answering. We also plan to improve our models with a phrase-based decoding procedure.


\bibliography{refs}
\bibliographystyle{iclr2019_conference}

\newpage
\input{supp.tex}

\end{document}

%% file: math_commands.tex
\usepackage{amsmath,amsfonts,bm}

\def\eqref#1{equation~\ref{#1}}

\def\1{\bm{1}}

\def\vzero{{\bm{0}}}

\def\vx{{\bm{x}}}
\def\vy{{\bm{y}}}
\def\vz{{\bm{z}}}

\def\mA{{\bm{A}}}

\def\mI{{\bm{I}}}

\def\mK{{\bm{K}}}

\def\mO{{\bm{O}}}

\def\mQ{{\bm{Q}}}

\def\mV{{\bm{V}}}
\def\mW{{\bm{W}}}
\def\mX{{\bm{X}}}

\def\mZ{{\bm{Z}}}

\DeclareMathAlphabet{\mathsfit}{\encodingdefault}{\sfdefault}{m}{sl}
\SetMathAlphabet{\mathsfit}{bold}{\encodingdefault}{\sfdefault}{bx}{n}

\newcommand{\softmax}{\mathcal{S}}

\DeclareMathOperator{\real}{\rm I\!R}

%% file: our-macro.tex
\usepackage{xspace}

\newcommand{\red}[1]{\textcolor{red}{#1}}

\newcommand{\ie}{{\em i.e.,}\xspace}
\newcommand{\eg}{{\em e.g.,}\xspace}

\newcommand{\aka}{\emph{a.k.a.}\xspace}


%% file: introduction.tex
\textbf{Neural Machine Translation (NMT)} has established breakthroughs in many different translation tasks, and has quickly become the standard approach to machine translation. 
NMT offers a simple encoder-decoder architecture that is trained end-to-end. 
Most NMT models (except a few like \citep{neural_gpu_NIPS2016_6295} and \citep{towardsphrasenmthuang2017towards}) possess attention mechanisms to perform alignments of the target tokens to the source tokens. 
The attention module plays a role analogous to the word alignment model in Statistical Machine Translation or SMT \citep{Koehn:2010:SMT}. In fact, the transformer network introduced recently by \cite{vaswani2017attention} achieves state-of-the-art performance in both speed and BLEU scores \citep{bleu_score_papineni2002bleu} by using only attention modules. 

{On the other hand, phrasal interpretation is an important aspect for many language processing tasks, and forms the basis of Phrase-Based Machine Translation \citep{Koehn:2010:SMT}. Phrasal alignments \citep{Koehn:2003:SPT} can model one-to-one, one-to-many, many-to-one, and many-to-many relations between source and target tokens, and use local context for translation. They are also robust to non-compositional phrases. Despite the advantages, the concept of \textbf{phrasal attentions} has largely been neglected in NMT, as most NMT models generate translations token-by-token autoregressively, and use the token-based attention method which is order invariant. Therefore, the intuition of phrase-based translation is vague in existing NMT systems that solely depend on the underlying neural architectures (recurrent, convolutional, or self-attention) to incorporate compositional information.}


In this paper, we propose phrase-based attention methods for phrase-level alignments in NMT. 
Specifically, we propose two novel phrase-based attentions, namely {\sc{\textbf{ConvKV}}} and {\sc{\textbf{QueryK}}}, designed to assign attention scores directly to phrases in the source and compute phrase-level attention vector for the target. We also introduce three new attention structures, which apply these methods to conduct phrasal alignments. Our \textbf{homogeneous} and \textbf{heterogeneous} attention structures perform \textit{token-to-token} and \textit{token-to-phrase} mappings, while the \textbf{interleaved heterogeneous} attention structure models all \textit{token-to-token, token-to-phrase, phrase-to-token}, and \textit{phrase-to-phrase} alignments.

To show the effectiveness of our approach, we apply our phrase-based attentions to all multi-head attention layers in the Transformer network. 
Our experiments on WMT'14 translation tasks between English and German show 1.3 BLEU improvement for English-to-German and 0.5 BLEU for German-to-English, 
compared to the Transformer network trained in identical settings.

%% file: background.tex
Most NMT models adopt an encoder-decoder framework, where the {encoder} network first transforms an input sequence of symbols $\vx = (x_1, x_2 \ldots x_n)$ to a sequence of continuous representations $\mZ = (\vz_1, \vz_2, \ldots \vz_n)$, from which the {decoder} generates a target sequence of symbols $\vy = (y_1, y_2, \ldots y_n)$ autoregressively, one element at a time. Recurrent seq2seq models with diverse structures and complexity \citep{quocle_seq2seq_NIPS2014_5346, bahdanau2014neural, luong2015effective, wu2016google} are the first to yield state-of-the-art results. Convolutional seq2seq models \citep{byte_net_kalchbrenner2016neural,gehring2017convolutional,slicenet_kaiser2018depthwise} alleviate the drawback of sequential computation of recurrent models and leverage parallel computation to reduce training time. 

The recently proposed \textbf{Transformer} network \citep{vaswani2017attention} structures 
the encoder and the decoder entirely with stacked self-attentions and cross-attentions (only in the decoder). In particular, it uses a multi-headed, scaled multiplicative attention defined as follows: 

\vspace{-1em}
\begin{eqnarray}
\text{Attention}(\mQ, \mK, \mV, \mW_q, \mW_k, \mW_v) \hspace*{-0.5em}&=& \hspace*{-0.5em}\softmax(\frac{(\mQ\mW_q) (\mK\mW_k)^T}{\sqrt{d_k}}) (\mV\mW_v) \\
\text{Head}^i \hspace*{-0.5em}&=& \hspace*{-0.5em}\text{Attention}(\mQ, \mK, \mV, \mW_q^i, \mW_k^i, \mW_v^i) ~~ \text{for}~~i = 1 \ldots h \\
\text{AttentionOutput}(\mQ, \mK, \mV, \mW) \hspace*{-0.5em}&=& \hspace*{-0.5em}\text{concat}(\text{Head}^1, \text{Head}^2, \ldots ,\text{Head}^h)\mW
\end{eqnarray}

where $\softmax$ is the \textit{softmax} function, $\mQ$, $\mK$, $\mV$ are the matrices with query, key, and value vectors, respectively, $d_k$ is the dimension of the query/key vectors; $\mW_q^i$, $\mW_k^i$, $\mW_v^i$ are the head-specific weights for query, key, and value vectors, respectively; and $\mW$ is the weight matrix that combines the outputs of the heads. The attentions in the {encoder} and {decoder} are based on \textbf{self-attention}, where all of $\mQ$, $\mK$ and $\mV$ come from the output of the previous layer. The decoder also has \textbf{cross-attention}, where $\mQ$ comes from the previous decoder layer, and the $\mK$-$\mV$ pairs come from the encoder. We refer readers to \citep{vaswani2017attention} for further details of the network design.

One crucial issue with the attention mechanisms employed in the Transformer network as well as other NMT architectures \citep{luong2015effective,gehring2017convolutional} is that they are  order invariant locally and globally.
If this problem is not tackled properly, the model may not learn the sequential characteristics of the data. RNN-based models \citep{bahdanau2014neural,luong2015effective} tackle this issue with a recurrent encoder and decoder, CNN-based models like \citep{gehring2017convolutional} use position embeddings, while the Transformer uses positional encoding. Another limitation is that these attention methods attend to tokens, and play a role analogous to word alignment models in SMT. It is, however, well admitted in SMT that phrases are better than words as translation units \citep{Koehn:2010:SMT}. Without specific attention to phrases, a particular attention function has to depend entirely on the token-level \textit{softmax} scores of a phrase for phrasal alignment, which is not robust and reliable, thus making it more difficult to learn the mappings.  

There exists some research on phrase-based decoding in NMT framework. For example, \cite{towardsphrasenmthuang2017towards} proposed a phrase-based decoding approach based on a soft reordering layer and a Sleep-WAke Network (SWAN), a segmentation-based sequence model proposed by \cite{wang2017SWAN}. Their decoder uses a recurrent architecture without any attention on the source.  \cite{externalphrasememtang2016neural} and \cite{translatephrasewang2017translating} used an external phrase memory to decode phrases for a Chinese-to-English translation task. In addition, hybrid search and PBMT were introduced to perform phrasal translation in \citep{Dahlmann-17-Phrase}. Nevertheless, to the best of our knowledge, our work is the first to embed phrases into attention modules, which thus propagate the information throughout the entire end-to-end Transformer network, including the encoder, decoder, and the cross-attention. 




%% file: phrase_based_attention.tex
In this section, we present our proposed methods to compute attention weights and vectors based on n-grams of queries, keys, and values. We compare and discuss the pros and cons of these methods. For simplicity, we describe them in the context of the Transformer network; however, it is straight-forward to apply them to other architectures such as RNN-based or CNN-based seq2seq models.

\subsection{Phrase-Based Attention Methods} \label{subsec:phattn}

In this subsection, we present two novel methods to achieve phrasal attention. In Subsection \ref{subsec:multi-head}, we present our methods for combining different types of n-gram attentions. 
The key element in our methods is a temporal (or one-dimensional) convolutional operation that is applied to a sequence of vectors representing tokens. Formally, we can define the \textbf{convolutional} operator applied to each token $x_t$ with corresponding vector representation $\mathbf{x}_t \in \real^{d_1}$ as:

\vspace{-1em}
\begin{eqnarray}
    o_t =  \mathbf{w} \oplus_{k=0}^n \mathbf{x}_{t \pm k} \label{eqn:conv}
\end{eqnarray}
\vspace{-1em}

where $\oplus$ denotes vector concatenation, $\mathbf{w} \in \real^{n \times d_1}$ is the weight vector (\aka kernel), and $n$ is the window size. We repeat this process with $d_2$ different weight vectors to get a $d_2$-dimensional latent representation for each token $x_t$. We will use the notation $\text{Conv}_n(\mX, \mW)$ to denote the convolution operation over an input sequence $\mX$ with window size $n$ and kernel weights $\mW \in \real^{n \times d_1 \times d_2}$.


\subsubsection{Key-Value Convolution}

The intuition behind \textbf{key-value convolution} technique is to use trainable kernel parameters $\mW_k$ and $\mW_v$ to compute the latent representation of n-gram sequences using convolution operation over key and value vectors. The attention function with key-value convolution is defined as:


\vspace{-1em}
\begin{eqnarray}
    \text{{\sc{ConvKV}}}(\mQ, \mK, \mV) =  \softmax(\frac{(\mQ\mW_q)\text{Conv}_n(\mK, \mW_k)^T}{\sqrt{d_k}})~\text{Conv}_n(\mV, \mW_v) \label{eqn:convkv}
\end{eqnarray}
\vspace{-1em}
        
        
\noindent where $\softmax$ is the \textit{softmax} function, $\mW_q \in{\real^{d_q \times d_k}}, \mW_k \in{\real^{n \times d_k \times d_k}}, \mW_v \in{\real^{n \times d_v \times d_v}}$ are the respective kernel weights for $\mQ$, $\mK$ and $\mV$. Throughout this paper, we will use $\softmax$ to denote the \textit{softmax} function. Note that in this convolution, the key and value sequences are left zero-padded so that the sequence length is preserved after the convolution (\ie\ one latent representation per token).
        
         
{The {\sc{ConvKV}} method can be interpreted as \emph{indirect} query-key attention, in contrast to the \emph{direct} query-key approach to be described next. This means that the queries do not interact directly with the keys to learn the attention weights; instead the model relies on the kernel weights ($\mW_k$) to learn n-gram patterns.}

 \subsubsection{Query-as-Kernel Convolution}

In order to allow the queries to \emph{directly} and \emph{dynamically} influence the word order of phrasal keys and values, we introduce \textbf{Query-as-Kernel} attention method. In this approach, when computing the attention weights, we use the query as kernel parameters in the convolution applied to the series of keys. The attention output in this approach is given by:

        
\hspace{-1.5em}
\begin{eqnarray}
\text{\sc{QueryK}}(\mQ, \mK, \mV) = \softmax(\frac{\text{Conv}_n(\mK\mW_k, \mQ\mW_q)}{\sqrt{d_k*n}})~\text{Conv}_n(\mV, \mW_v) \label{eqn:queryk}
\end{eqnarray} 
\hspace{-1.5em}
        
where $\mW_q \in{\real^{n \times d_q \times d_k}}, \mW_k \in{\real^{d_k \times d_k}}, \mW_v \in{\real^{n \times d_v \times d_v}}$ are trainable weights. Notice that we include the window size $n$ (phrase length) in the scaling factor to counteract the fact that there are $n$ times more multiplicative operations in the convolution than the traditional matrix multiplication. 


\subsection{Multi-Headed Phrasal Attention}  
\label{subsec:multi-head}

We now present our extensions to the multi-headed attention framework of the Transformer to enable it to pay attention not only to tokens but also to phrases across many sub-spaces and locations.      




\subsubsection{Homogeneous N-Gram Attention}  

\begin{figure}[t!]
    \vspace{-1em}
    \begin{center}
        \includegraphics[width=0.7\textwidth]{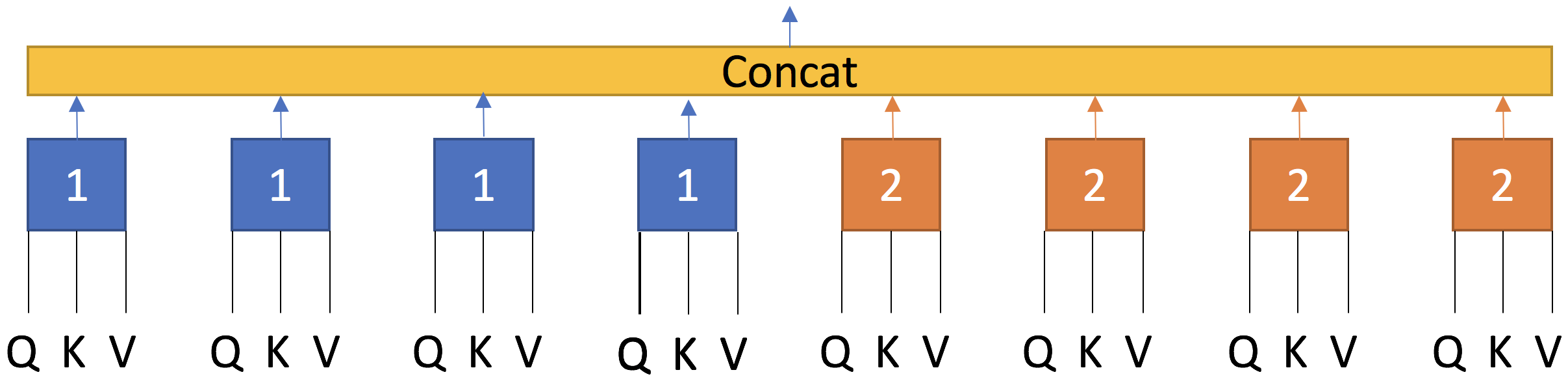}
    \end{center}
    \vspace{-1em}
    \caption{Homogeneous multi-head attention, where each attention head features one n-gram type. In this example, there are eight heads, which are distributed equally between unigrams and bigrams.}
    \label{fig:homo}
    \end{figure}

In \textbf{homogeneous n-gram attention}, we distribute the heads to different n-gram types with each head attending to one particular n-gram type (n=$1,2,\ldots, N$). For instance,  Figure \ref{fig:homo} shows a homogeneous structure, where the first four heads attend to unigrams, and the last four attend to bigrams. A head can apply one of the phrasal attention methods described in Subsection \ref{subsec:phattn}. The selection of which n-gram to assign to how many heads is arbitrary. Since all heads must have consistent sequence length, phrasal attention heads require left-padding of keys and values before convolution.
     
     Since each head attends to a subspace resulting from one type of n-gram, homogeneous attention learns the mappings in a distributed way. However, the homogeneity restriction may limit the model to learn interactions between different n-gram types. Furthermore, the homogeneous heads force the model to assign each query with attentions on all n-gram types (\eg\ unigrams and bigrams) even when it does not need to do so, thus possibly inducing more noise into the model.


     
    
\subsubsection{Heterogeneous N-Gram Attention}

\begin{figure}[t!]
    \vspace{-1em}
    \begin{center}
        \includegraphics[width=\textwidth]{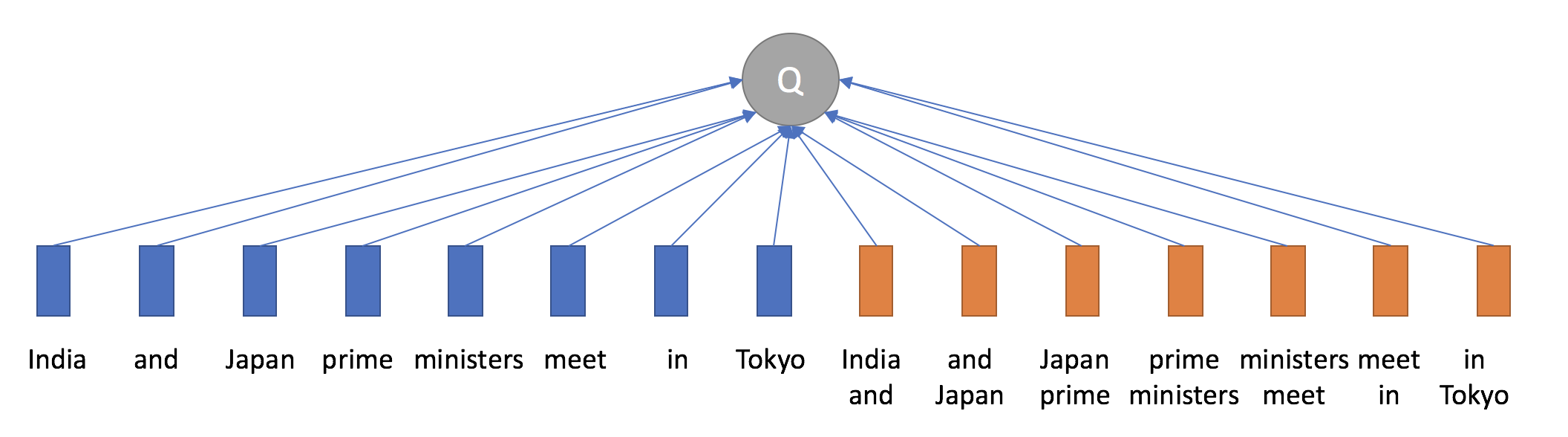}
    \end{center}
    \vspace{-1.2em}
    \caption{Heterogeneous n-gram attention for each attention head. Attention weights and vectors are computed from all n-gram types simultaneously.}
    \label{fig:hete}
    \end{figure}

The \textbf{heterogeneous n-gram attention} relaxes the constraint of the homogeneous approach. Instead of limiting each head's attention to a particular type of n-gram, it allows the query to freely attend to all types of n-grams simultaneously. To achieve this, we first compute the attention logit for each n-gram type separately (\ie\ for n $=1, 2, \ldots, N$), then we concatenate all the logits before passing them through the \textit{softmax} layer to compute the attention weights over all n-gram types. Similarly, the value vectors for the n-gram types are concatenated to produce the overall attention output. Figure \ref{fig:hete} demonstrates the heterogeneous attention process for unigrams and bigrams.        
    
For {\sc{ConvKV}} technique in Equation \ref{eqn:convkv}, the attention output is given by:
    
\vspace{-1em}
\begin{eqnarray}
\softmax(\frac{(\mQ\mW_q)[(\mK\mW_{k,1})^T;\text{Conv}_{2}(\mK,\mW_{k,2})^T;...]}{\sqrt{d_k}})[(\mV\mW_{v,1}); \text{Conv}_{2}(\mV,\mW_{v,2});...] \label{eqn:hete_convkv}
\end{eqnarray}
\vspace{-1em}

For {\sc{QueryK}} technique (Equation \ref{eqn:queryk}), the attention output is given as follows:
    
    
\vspace{-1em}
\begin{eqnarray}
\softmax([\frac{(\mQ\mW_{q,1})(\mK\mW_{k,1})^T}{\sqrt{d}};\frac{\text{Conv}_{2}(\mK\mW_{k,2},\mQ\mW_{q,2})}{\sqrt{d*n_2}};...]) [(\mV\mW_{v,1});\text{Conv}_{2}(\mV,\mW_{v,2});...] \label{eqn:hete_queryk}
\end{eqnarray}
\vspace{-1em}

Note that in heterogeneous attention, we do not need to pad the input sequences before the convolution operation to ensure identical sequence length. Also, the key/value sequences that are shorter than the window size do not have any valid phrasal component to be attended.





\subsection{Interleaved Phrases to Phrase Heterogeneous Attention}

All the methods presented above perform attention mappings from token-based queries to phrase-based key-value pairs. In other words, they adopt token-to-token and token-to-phrase structures.
These types of attentions are beneficial when there exists a translation of a phrase in the source language (keys and values) to a single token in the target language (query). However, these methods are not explicitly designed to work in the reverse direction when a phrase or a token in the source language should be translated to a phrase in the target language. In this section, we present a novel approach to heterogeneous phrasal attention that allows phrases of queries to attend on tokens and phrases of keys and values (\ie\ phrase-to-token and phrase-to-phrase mappings). 


    
    
\begin{figure}[tb!]
\vspace{-1em}    
\begin{center}
\includegraphics[width=0.75\textwidth]{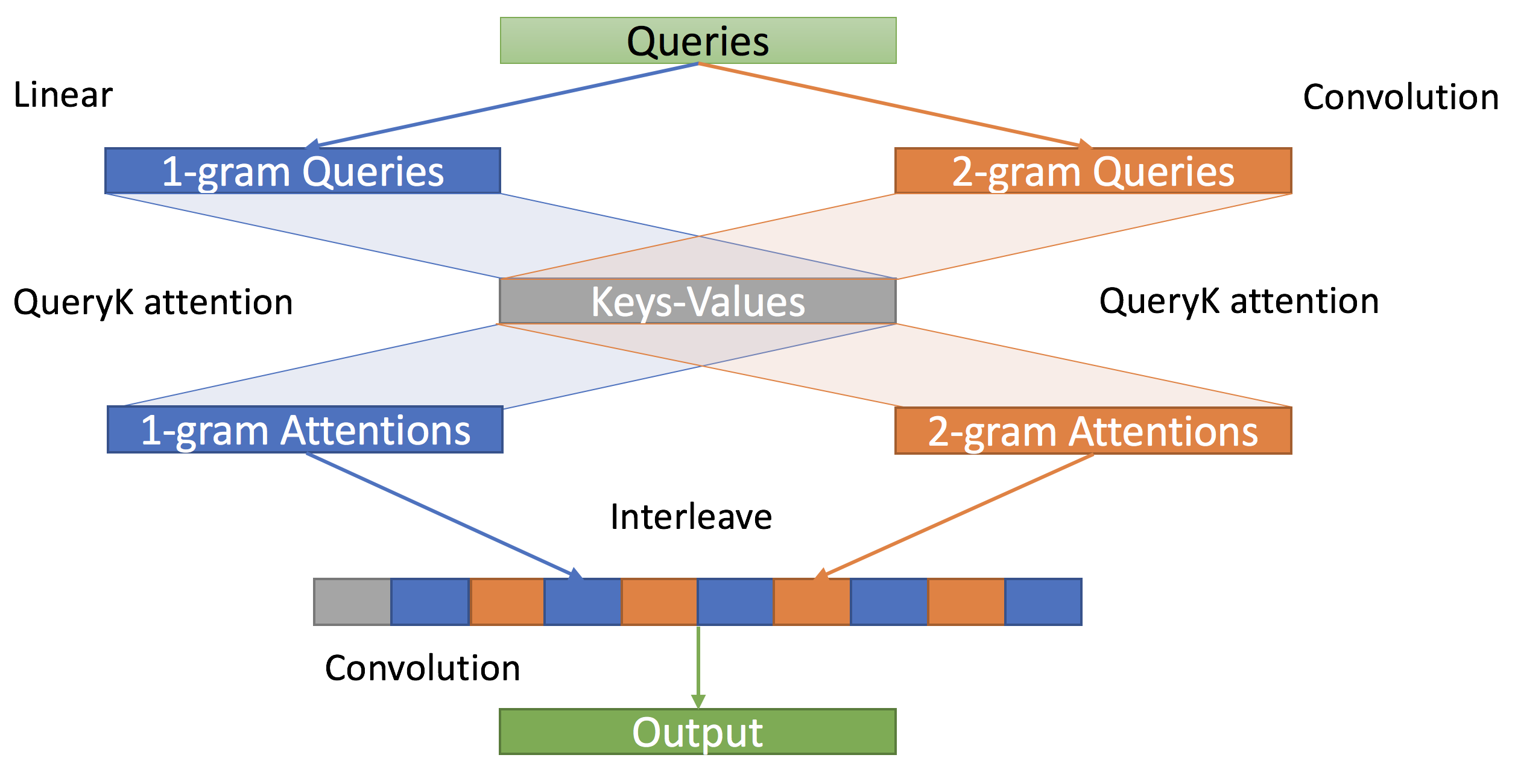}
\end{center}
\vspace{-1em}    
\caption{Interleaved phrase-to-phrase heterogeneous attention. The queries are first transformed into unigram and bigram representations, which in turn then attend independently on key-value pairs to produce unigram and bigram attention vectors. The attention vectors are then interleaved before passing through another convolutional layer.}
\label{fig:interleave}
\end{figure}

We accomplish this with the {\sc{QueryK}} and {\sc{ConvKV}} methods as follows. We first apply convolutions $\text{Conv}_n(\mQ, \mW_{q_n})$ on the query sequence with kernel weights $\mW_{q_n}$ for window size $n$ to obtain the hidden representations for $n$-grams. Consider  Figure \ref{fig:interleave}, where we apply convolution on $\mQ$ for $n=1$ and $n=2$ to generate the respective unigram and bigram queries.\footnote{$\text{Conv}_1(\mQ, \mW_{q_1})$ is equivalent to a feed-forward connection.} These queries are then used to attend over unigram and bigram key-values to generate the heterogeneous attention vectors. The result of these operations is a sequence of unigram and bigram attention vectors $\mA_1 = (\mathbf{u}_1, \mathbf{u}_2, \ldots, \mathbf{u}_N)$ and $\mA_2 = (\mathbf{b}_1, \mathbf{b}_2, \ldots, \mathbf{b}_{N-1})$ respectively, where $N$ is the query length.

    


\small
\vspace{-1em}
\begin{eqnarray}
\mA_{1,\text{ConvKV}}=& \hspace*{-3cm} \softmax(\frac{(\mQ\mW_{q_1})[(\mK\mW_{k,1})^T;\text{Conv}_{2}(\mK,\mW_{k,2})^T]}{\sqrt{d_k}}) [(\mV\mW_{v,1});\text{Conv}_{2}(\mV,\mW_{v,2})]  \\
\mA_{2,\text{ConvKV}}=& \hspace*{-2.3cm}\softmax(\frac{\text{Conv}_2(\mQ,\mW_{q_2})[(\mK\mW_{k,1})^T;\text{Conv}_{2}(\mK,\mW_{k,2})^T]}{\sqrt{d_k}})[(\mV\mW_{v,1});\text{Conv}_{2}(\mV,\mW_{v,2})]  \\
\mA_{1,\text{QueryK}}=& \hspace*{-1.8cm} \softmax([\frac{(\mQ\mW_{q_1,1})(\mK\mW_{k,1})^T}{\sqrt{d}};\frac{\text{Conv}_{2}(\mK\mW_{k,2},\mQ\mW_{q_1,2})}{\sqrt{d*n_2}}])[(\mV\mW_{v,1});\text{Conv}_{2}(\mV,\mW_{v,2})]  \\
\mA_{2,\text{QueryK}}=& \hspace*{-0.2cm}\softmax([\frac{\text{Conv}_2(\mQ,\mW_{{q}_2,1})(\mK\mW_{k,1})^T}{\sqrt{d}};\frac{\text{Conv}_{2}(\mK\mW_{k,2},\text{Conv}_2(\mQ,\mW_{q_2,2}))}{\sqrt{d*n_2}}])[(\mV\mW_{v,1});\text{Conv}_{2}(\mV,\mW_{v,2})]
\vspace{-0.3em}
\end{eqnarray}
\normalsize



Notice that each $\mathbf{b}_i \in \mA_2$ represents the attention vector for $(\mQ_{i}$-$\mQ_{i+1})$ bigram queries. In the next step, the phrase-level attention states in $\mA_2$ are \textit{interleaved} with the unigram attentions in $\mA_1$ to form an interleaved attention sequence $\mI$. For unigram and bigram queries, the interleaved vector sequences at the encoder and decoder are formed as

\vspace{-1em}    
\begin{eqnarray}
\mI_{\text{enc}} &=& (\vzero, \mathbf{u}_1, \mathbf{b}_1, \mathbf{u}_2, \mathbf{b}_2,\mathbf{u}_3, \ldots, \mathbf{b}_{N-1}, \mathbf{u}_N, \vzero)\\
\{ \mI_{\text{dec}}, {\mI_{\text{cross}}}\} &=& (\vzero, \mathbf{u}_1, \mathbf{b}_1, \mathbf{u}_2, \mathbf{b}_2, \mathbf{u}_3, \ldots, \mathbf{b}_{N-1}, \mathbf{u}_N)
\end{eqnarray}
\vspace{-1em}    

where $\mI_{\text{enc}}$ and $\mI_{\text{dec}}$ denote the interleaved sequence for self-attention at the encoder and decoder respectively, and $\mI_{\text{cross}}$ denotes the interleaved sequence for cross-attention between the encoder and the decoder. {Note that, to prevent information flow from the future in the decoder, the right connections are masked out in $\mI_{\text{dec}}$ and  ${\mI_{\text{cross}}}$ (similar to the original Transformer).} The interleaving operation places the phrase- and token-based representations of a token next to each other.  The interleaved vectors are passed through a convolution layer (as opposed to a point-wise feed-forward layer in the Transformer) to compute the overall representation for each token. By doing so, each query is intertwined with the n-gram representations of the phrases containing itself,  which enables the model to learn the query's correlation with neighboring tokens. For unigram and bigram queries, the encoder uses a convolution layer with a window size of 3 and stride of 2 to allow the token to intertwine with its past and future phrase representations, while the ones in the decoder (self-attention and cross-attention) use a window size of 2 and stride of 2 to incorporate only the past phrase representations to preserve the autoregressive property. More formally,  




\vspace{-1.5em}
\begin{eqnarray}
        \mO_{\text{enc}} &=& \text{Conv}_{\text{window=3,stride=2}}(\mI_{\text{enc}}, \mW_{\text{enc}})\\
        \mO_{\text{cross}} &=& \text{Conv}_{\text{window=2,stride=2}}(\mI_{\text{cross}}, \mW_{\text{cross}}) \\
        \mO_{\text{dec}} &=& \text{Conv}_{\text{window=2,stride=2}}(\mI_{\text{dec}}, \mW_{\text{dec}})
\end{eqnarray}
\vspace{-1.5em}

%% file: experiments.tex
\vspace{-0.5em}
In this section, we present the training settings, experimental results and analysis of our models.

\subsection{Training Settings}

We preserve most of the training settings from \cite{vaswani2017attention} to enable a fair comparison with the original Transformer. Specifically, we use the Adam optimizer \citep{kingma2014adam} with $\beta_1=0.9$, $\beta_2=0.98$, and $\epsilon=10^{-9}$. We follow a similar learning rate schedule with $warmup\_steps$ of $16000$:     $LearningRate = 2 * d^{-0.5} * \min(step\_num^{-0.5}, step\_num * warmup\_steps^{-1.5})$.


We trained our models and the baseline on a single GPU for 500,000 steps.\footnote{We were unable to replicate state-of-the-art results  in \citep{vaswani2017attention} because of limited GPUs. Hence, we conducted all the experiments including Transformer base in an identical setup for fair comparisons.} The batches were formed by sentence pairs containing approximately 4096 source and 4096 target tokens. Similar to \cite{vaswani2017attention}, we also applied residual dropout with 0.1 probability and label smoothing with $\epsilon_{ls} = 0.1$. Our models are implemented in the  \textit{tensor2tensor}\footnote{https://github.com/tensorflow/tensor2tensor} library \citep{tensor2tensor}, on top of the original Transformer codebase.  We trained our models on the standard WMT'16 English-German dataset constaining about 4.5 million sentence pairs, using WMT \textbf{newstest2013} as our development set and \textbf{newstest2014} as our test set.  
We used byte-pair encoding \citep{sennrich2015neural} with combined source and target vocabulary of 37,000 sub-words for English-German. We took the average of the last 5 checkpoints (saved at 10,000-iteration intervals) for evaluation, and used a beam search size of 5 and length penalty of 0.6 \citep{wu2016google}.

\subsection{Results}

Table \ref{table:bleu-en-de} compares our model variants with the Transformer base model of \cite{vaswani2017attention} on the \textbf{newstest2014} test set. All models were trained on identical settings. We notice that all of our models achieve higher BLEU scores than the baseline, showing the effectiveness of our approach.




On the \textbf{Engligh-to-German ({En-De})}  translation task, our \textbf{homogeneous} structure using {\sc{ConvKV}} phrasal attention with 44 head distribution already outperforms the Transformer base by more than 0.5 BLEU, while the one using {\sc{QueryK}} performs even better with a BLEU of 26.78. When we compare our \textbf{heterogeneous} models with homogeneous ones, we notice even higher scores for the heterogeneous models -- the {\sc{ConvKV}} achieves a score of 27.04, and the {\sc{QueryK}} achieves a score of 26.95. This shows the effectiveness of relaxing the `same n-gram type' attention constraint in the heterogeneous approach, which allows it to attend to all n-gram types simultaneously within a single head, avoiding any forced attention to a particular n-gram type. 
In fact, our experiments show that the models perform worse than the baseline if we remove token-level (unigram) attentions entirely.

\begin{table}[t]
\begin{center}
\begin{tabular}{llccc} 
\toprule
{\bf Model}                 & {\bf Technique}   & {\bf N-grams}  & {\bf En-De}      & {\bf De-En}        \\
\midrule
Transformer (Base, 1 GPU)          &-                  &-               &26.07             &29.82              \\
Transformer (Base, 8 GPUs)         &-                  &-               &27.30             &----              \\
\cite{vaswani2017attention}        &                   &                &                 &         	\\
\midrule
Homogeneous                 & {\sc{ConvKV}}     &44              &26.60 (+0.53)     &30.17 (+0.36)      \\
Homogeneous                 & {\sc{QueryK}}     &44              &26.78 (+0.71)     &30.03 (+0.21)      \\
\midrule
Heterogeneous               & {\sc{ConvKV}}     &12              &27.04 (+0.97)     &30.09 (+0.27)      \\
Heterogeneous               & {\sc{QueryK}}     &12              &26.95 (+0.88)     &30.20 (+0.38)      \\
\midrule
Interleaved                 & {\sc{ConvKV}}     &12              &27.33 (+1.26)     &30.17 (+0.36)      \\
Interleaved                 & {\sc{QueryK}}     &12              &\textbf{27.40 (+1.33)}   &\textbf{30.30 (+0.48)}      \\
\bottomrule
\end{tabular}
\vspace{-0.3em}
\caption{BLEU (cased) scores on WMT'14 testset for English-German and German-English. For homogeneous models, the \textbf{N-grams} column denotes how we distribute the 8 heads to different n-gram types;  \eg\ 323 means 3 unigram heads, 2 bigram heads and 3 trigram heads. For heterogeneous, the numbers indicate the phrase lengths of the collection of n-gram components jointly attended by each head; \eg\ 12 means attention scores are computed across unigram and bigram logits.}
\label{table:bleu-en-de}
\end{center}
\end{table}

\begin{table}[t]
\vspace{-1em}
\begin{center}
\begin{tabular}{llcccc} 
\toprule
{\bf Model}         & {\bf Technique}       & \multicolumn{2}{c}{\bf Uni-bi-grams}    & \multicolumn{2}{c}{\bf Uni-bi-tri-grams}     \\
{\bf      }         & {\bf          }           & 
{Head/N-gram} & {BLEU}         & {Head/N-gram} & {BLEU}   \\
\midrule
Homogeneous         & {\sc{ConvKV}} & 44            &26.60             & 323     &26.55              \\
Homogeneous         & {\sc{QueryK}}  & 44           &26.78              & 323    &26.86              \\
\midrule
Heterogeneous       & {\sc{ConvKV}}  & 12           &27.04              &123    &27.15              \\
Heterogeneous       & {\sc{QueryK}}  & 12            &26.95             &123     &27.09              \\
\bottomrule
\end{tabular}
\vspace{-0.3em}
\caption{BLEU scores for models that use only uni-bi-grams vs. the ones that use uni-bi-tri-grams.}
\label{table:n-gram-analysis}
\end{center}
\end{table}

Finally, we notice that our \textbf{interleaved} heterogeneous models surpass all aforementioned scores achieving up to 27.40 BLEU and establishing a 1.33 BLEU improvement over the Transformer. This demonstrates the existence of phrase-to-token and phrase-to-phrase mappings from target to source language within their mutual latent space. This is especially necessary when the target language is morphologically rich, like German, whose words are usually compounded with sub-words expressing different meanings and grammatical structures. Phrase-to-phrase mapping helps model local agreement, \eg\ between an adjective and a noun (in terms of gender, number and case) or between subject and verb (in terms of person and number). Our interleaved models with BPE \citep{sennrich2015neural} tackle all four possible types of alignments (token-to-token, token-to-phrase, phrase-to-token, phrase-to-phase) to counteract such linguistic differences in morphology and syntax.       

   

On \textbf{German-to-English ({De-En})} translation, likewise, all of our models achieve improvement compared to the Transformer base, but the gain is not as high as in the English-to-German task. Specifically, homogeneous and heterogeneous attentions perform comparably, giving up to +0.38 BLEU improvement compared to the Transformer base. Our interleaved models show more improvements (up to 30.30 BLEU), outperforming the Transformer by about 0.5 points. This demonstrates the importance of phrase-level query representation in the target.




Considering \textbf{\sc{ConvKV}} vs. {\sc{QueryK}}, both methods perform rather equivalently. {\sc{QueryK}} is better than {\sc{ConvKV}} for homogeneous models but worse for their heterogeneous counterparts on the English-to-German task. However, the opposite trend can also be observed for German-to-English in one case.
Meanwhile, {\sc{QueryK}} achieves higher BLEU in interleaved models in both  tasks.

For further analysis, Table \ref{table:n-gram-analysis} shows how the performance differs if we include a trigram component in our homogeneous and heterogeneous models.\footnote{It is nontrivial to use unigram, bigram and trigram components in the interleaved model, because one needs to mix three sequences into one such that different n-gram types align properly. We leave it to future work.} For homogeneous models, {\sc{ConvKV}} leads to a deterioration in performance, while {\sc{QueryK}} leads to a minor improvement. By contrast, heterogeneous models achieve higher BLEU scores for both techniques. This supports the argument that some trigrams are not useful, and heterogeneous models allow the queries to select any type of phrasal keys and values, whereas homogeneous ones force them to attend to unnecessary n-grams, inducing more noise.

\begin{figure}
\vspace{-1em}
	\centering
	\begin{subfigure}[c]{0.45\textwidth}
	    \centering
		\includegraphics[width=0.9\textwidth]{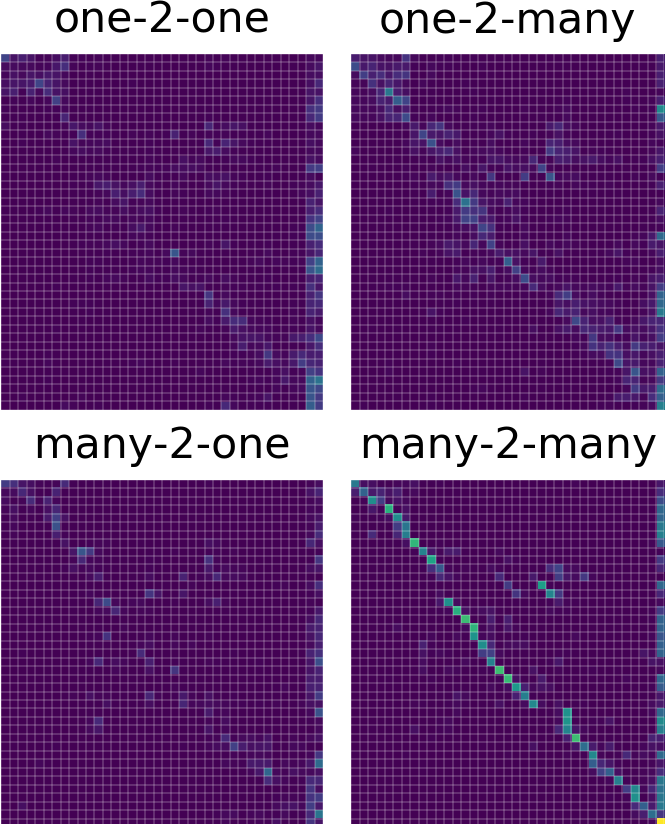}
		\caption{Layer 3} 
		\label{fig:att_layer_3}
	\end{subfigure}
    \hspace{2em}
	\vspace{1em} 
	\begin{subfigure}[c]{0.45\textwidth}
	    \centering
		\includegraphics[width=0.9\textwidth]{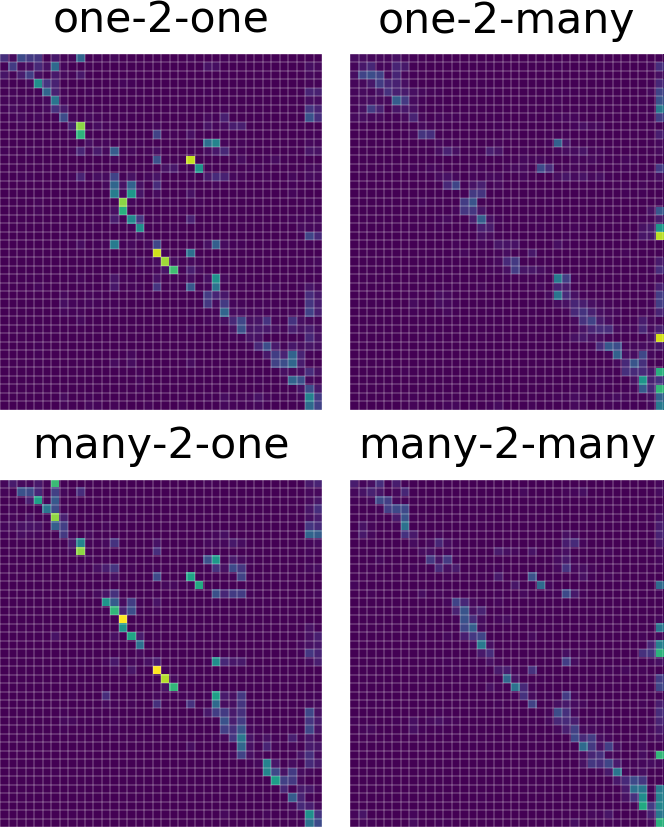}
		\caption{Layer 6} 
		\label{fig:att_layer_6}
	\end{subfigure}
    \vspace{-1em}
	\caption{Sample attention heat map of {\sc{QueryK}} interleaved heterogeneous model.} 
	\label{fig:attention_map}
\end{figure}

To interpret our phrasal attention models, we now discuss how they learn the alignments. Figure \ref{fig:attention_map} shows the attention heat maps for an English-German sample in {newstest2014}; figure \ref{fig:att_layer_3} displays the heat map in layer 3 (mid layer) while figure \ref{fig:att_layer_6} shows the one in layer 6 (top layer) within a six-layer transformer model based on our interleaved attention. Each figure shows 4 quadrants representing \textit{token-to-token, token-to-phrase, phrase-to-token, phrase-to-phrase} attentions, respectively. It can be seen that phrasal attentions are activated strongly in the mid-layers; particularly, the phrase-to-phrase attention is the most concentrated. The models learn phrasal alignments in the middle of the network. On the other hand, token-to-token attention is activated the most in the top layer, which is understandable because the top layer connects directly with the peripheral \textit{softmax} layer to generate translations token-by-token autoregressively. In fact, we observed that phrasal attention intensity transits gradually from phrase-level attention in the lower layer to token-level attention in the upper layer, and this phenomenon occurs frequently. Heterogeneous models follow the same trend as well.

Table \ref{table:att_percent} presents quantitatively the activation percentage of each attention type per layer computed by averaging over all sentences in the English-to-German newstest2014. We notice that attentions are distributed unevenly across four alignment types and differently in each layer. Phrase-to-phrase attention interestingly accounts for 96.16\% in layer 3, while layers 5 and 6 concentrate attention scores mostly in token-to-token mappings \footnote{Note that the layer 1 also squeezes attention scores to the token-to-token quadrant. However, it does not learn the proper token mappings; instead it concentrates the weights to the last \textit{$<$eos$>$} token.}. This indicates a particular combination of mapping types is learned distinctively by a specific layer of the network. We refer the readers to the Appendix (Section \ref{appendix}) for more figures and details about attention heat maps and statistics.


\begin{table}[t]
\vspace{-0.5em}
\begin{center}
\resizebox{0.8\columnwidth}{!}{%
\begin{tabular}{ccccc} 
\toprule
{\bf Layer}         & {\bf token-to-token}       & {\bf token-to-phrase}     & {\bf phrase-to-token}   & {\bf phrase-to-phrase} \\
\midrule
1                   &98.01                       &00.20                      &01.60                    &00.19                   \\
2                   &39.63                       &19.49                      &02.73                    &38.15                   \\
3                   &01.54                       &02.30                      &00.00                    &96.16                   \\
4                   &36.13                       &37.53                      &06.60                    &19.74                   \\
5                   &61.96                       &18.12                      &08.59                    &11.33                   \\
6                   &53.72                       &10.53                      &34.63                    &01.12                    \\
\bottomrule
\end{tabular}
}
\vspace{-0.3em}
\caption{Activation percentages for different attention types in each layer of the {Interleaved} model.} 

\label{table:att_percent}
\end{center}
\end{table}

%% file: supp.tex
\section{Appendix}\label{appendix}

\subsection{Attention Heat Map of QueryK Heterogeneous Model}

\begin{figure}[h!]
	\centering
	\begin{subfigure}[c]{1.0\textwidth}
	    \centering
		\caption{Layer 3 of {\sc{QueryK}} heterogeneous} 
		\includegraphics[width=\textwidth]{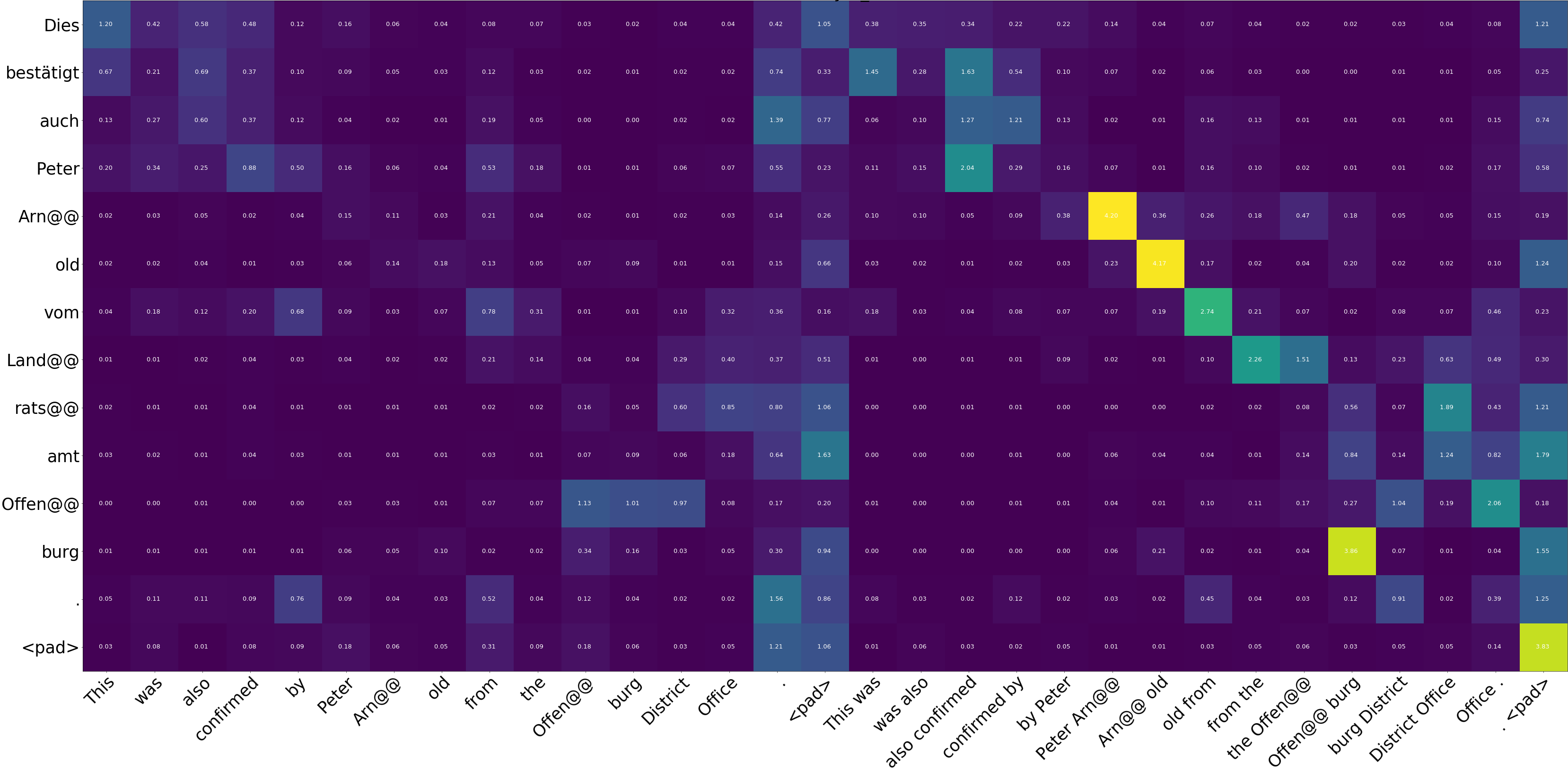}
		\label{fig:hete_att_layer_3}
	\end{subfigure}
	\vspace{1em} 
	\begin{subfigure}[c]{1.0\textwidth}
	    \centering
		\caption{Layer 6 {\sc{QueryK}} heterogeneous} 
		\includegraphics[width=\textwidth]{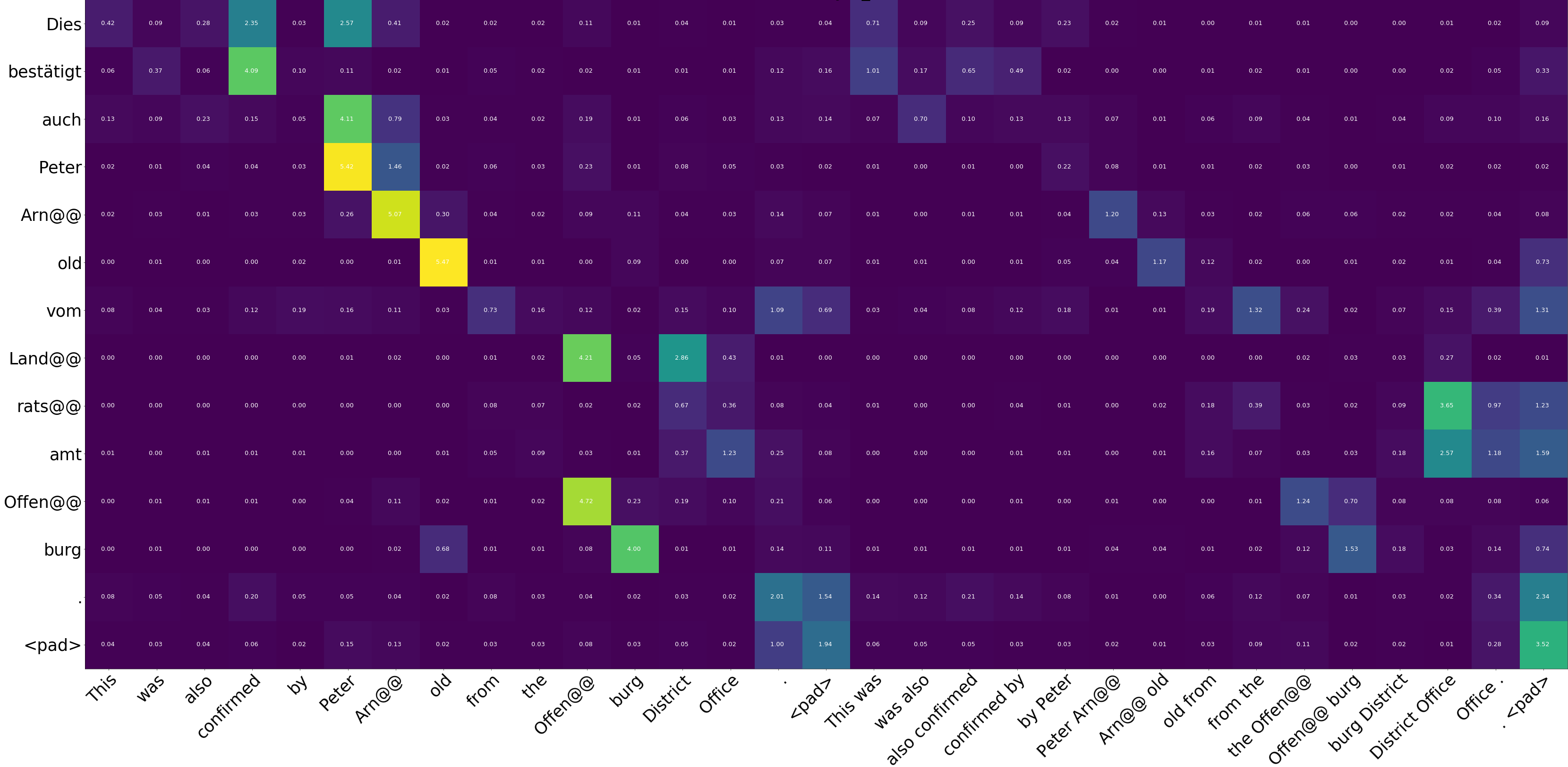}
		\label{fig:hete_att_layer_6}
	\end{subfigure}
    \vspace{-1em}
	\caption{Attention heat maps at layer 3 and layer 6 of {\sc{\textbf{QueryK}}} \textbf{heterogeneous} model for a sample sentence pair in English-German newstest2014 test set. The left half in each figure indicates \textbf{token-to-token} mappings, while the right half indicates \textbf{token-to-phrase} mappings.} 
	\label{fig:attention_map_hete}
\end{figure}

\subsection{Attention Statistics for QueryK Heterogeneous Model}

Table \ref{table:hete_att_percent} presents quantitatively the activation percentage of each attention type per layer for {\sc{QueryK}} heterogeneous model computed by averaging over all sentences in the English-to-German newstest2014. The mid layers (2, 3, 4, 5) activate mostly (if not entirely) phrasal attentions while only the first and last layer emphasize token-level attentions.


\begin{table}[h!]
\begin{center}
\begin{tabular}{ccc} 
\toprule
{\bf Layer}         & {\bf token-to-token}       & {\bf token-to-phrase}     \\
\midrule
1                   &100.00                      &00.00                      \\
2                   &00.24                       &99.76                      \\
3                   &01.10                       &98.90                      \\
4                   &05.83                       &94.17                      \\
5                   &23.84                       &76.16                      \\
6                   &86.30                       &13.70                      \\
\bottomrule
\end{tabular}
\caption{Activation percentages for different attention types in each layer of the {\sc{\textbf{QueryK}}} \textbf{heterogeneous attention} model. These numbers are computed by averaging over all sentences in newstest2014 for En-De translation task. The mid layers (2, 3, 4, 5) activate mostly (if not entirely) phrasal attentions while only the first and last layer emphasize token-level attentions.}
\label{table:hete_att_percent}
\end{center}
\end{table}


\newpage
\subsection{Attention Heat Map of QueryK Interleaved Model}

\begin{figure}[h!]
	\centering
	\begin{subfigure}[c]{0.95\textwidth}
	    \centering
		\caption{Layer 3 of {\sc{QueryK}} interleaved} 		\includegraphics[width=0.8\textwidth]{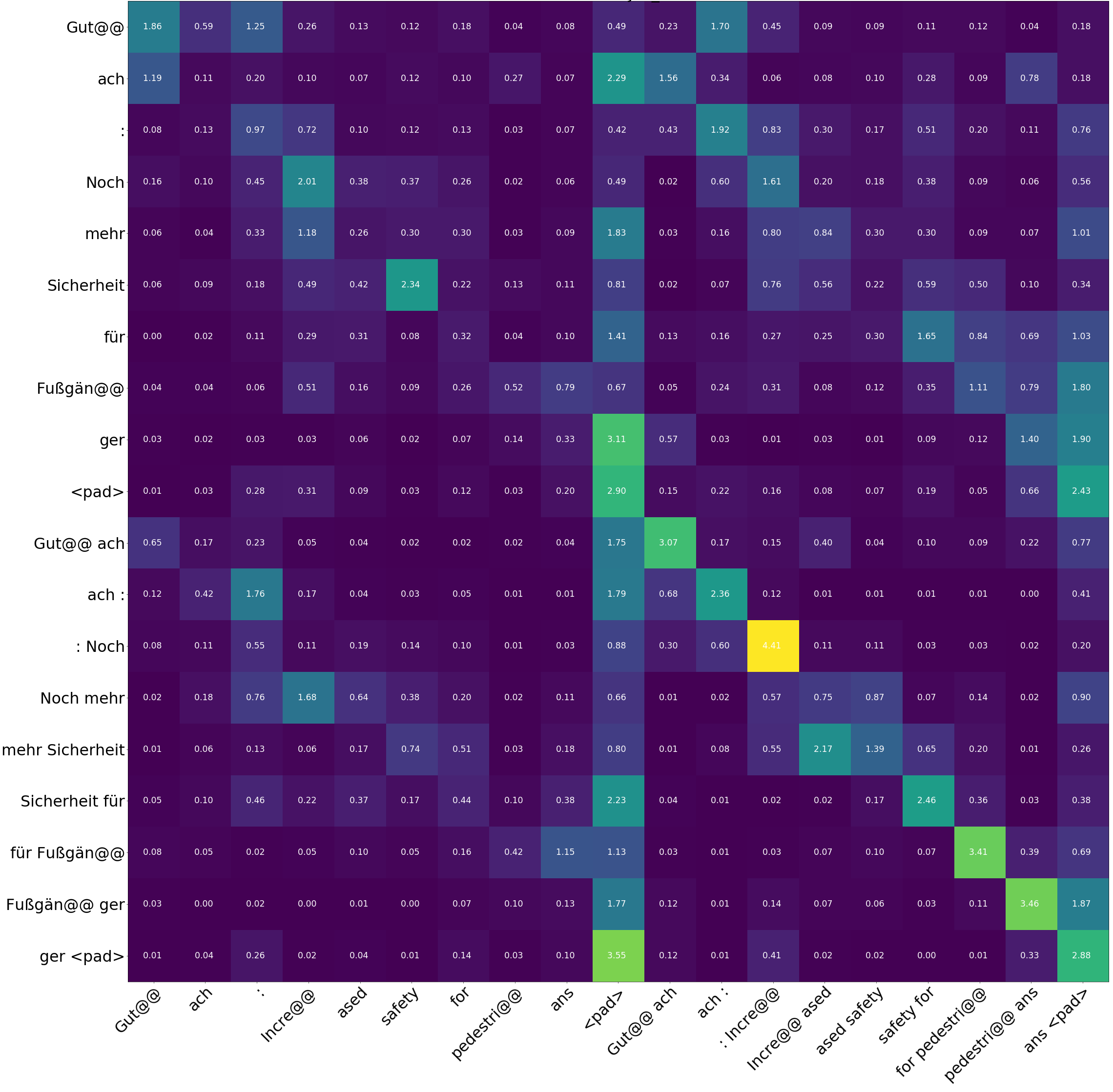}
		\label{fig:inter_0_att_layer_3}
	\end{subfigure}
    \hspace{2.5em}
	\vspace{1em} 
	\begin{subfigure}[c]{0.95\textwidth}
	    \centering
		\caption{Layer 6 {\sc{QueryK}} interleaved} 
		\includegraphics[width=0.8\textwidth]{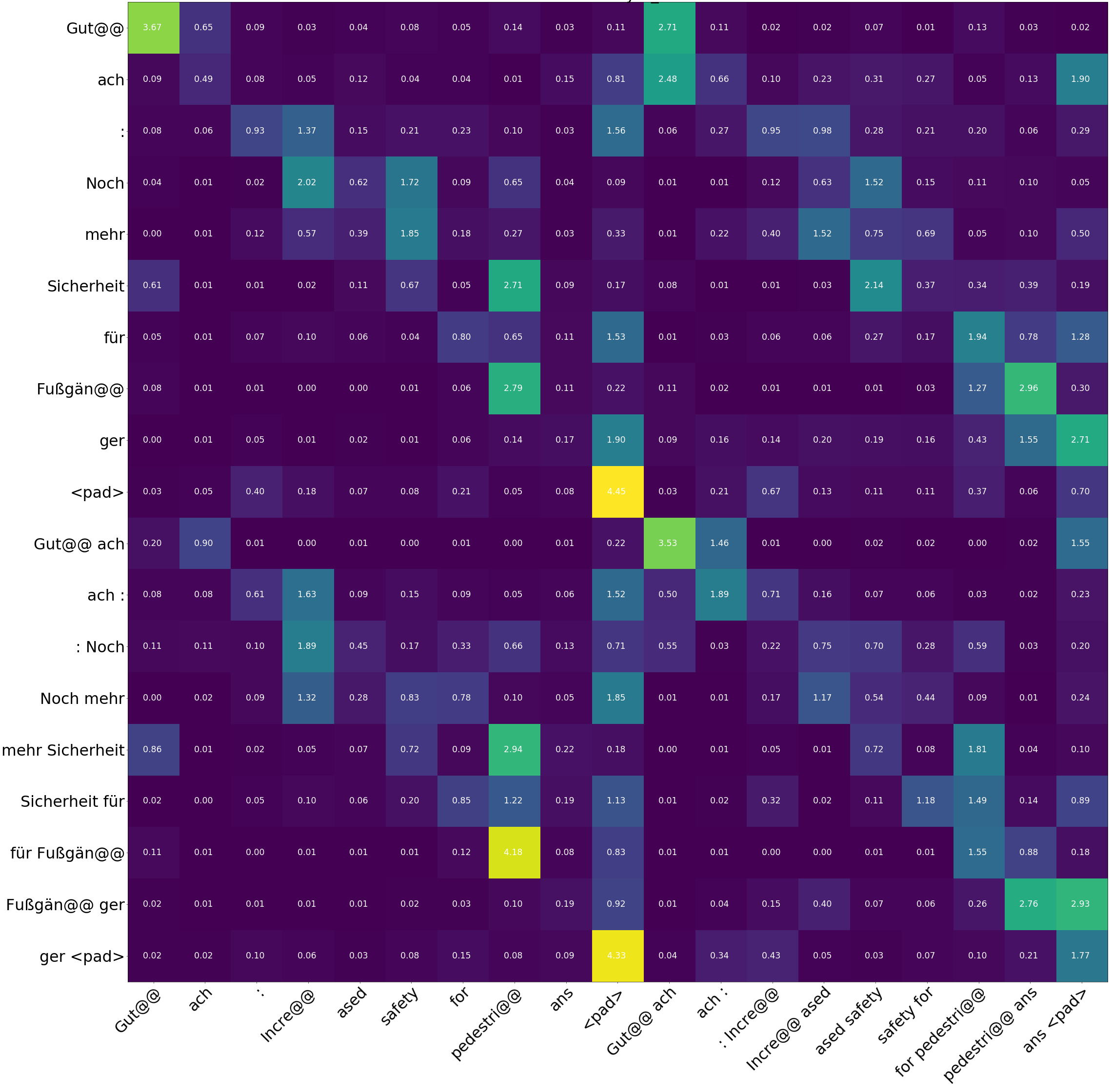}
		\label{fig:inter_0_att_layer_6}
	\end{subfigure}

	\caption{Attention heat maps at layer 3 and layer 6 of {\sc{\textbf{QueryK}}} \textbf{interleaved heterogeneous} model for another sample from English-German newstest2014 test set. Upper-left, upper-right, lower-left, lower-right quadrants of each figure show token-to-token, token-to-phase, phrase-to-token, phrase-to-phrase alignments respectively.} 

	\label{fig:attention_map_inter_0}
\end{figure}